# Multi-fidelity Design of Porous Microstructures for Thermofluidic Applications

Jonathan Tammer Eweis-Labolle, Chuanning Zhao, Yoonjin Won[†], and Ramin Bostanabad[†]

Department of Mechanical and Aerospace Engineering, University of California, Irvine

**Abstract**

As modern electronic devices are increasingly miniaturized and integrated, their performance relies more heavily on effective thermal management. Two-phase cooling methods enhanced by porous surfaces, which capitalize on thin-film evaporation atop structured porous surfaces, are emerging as potential solutions. In such porous structures, the optimum heat dissipation capacity relies on two competing objectives that depend on mass and heat transfer. The computational costs of evaluating these objectives, the high dimensionality of the design space which a voxelated microstructure representation, and the manufacturability constraints hinder the optimization process for thermal management. We address these challenges by developing a data-driven framework for designing optimal porous microstructures for cooling applications. In our framework we leverage spectral density functions (SDFs) to encode the design space via a handful of interpretable variables and, in turn, efficiently search it. We develop physics-based formulas to quantify the thermofluidic properties and feasibility of candidate designs via offline simulations. To decrease the reliance on expensive simulations, we generate multi-fidelity data and build emulators to find Pareto-optimal designs. We apply our approach to a canonical problem on evaporator wick design and obtain fin-like topologies in the optimal microstructures which are also characteristics often observed in industrial applications.

## 1 Introduction

Developments in modern electronic devices such as smartphones, computers, and medical devices [1] heavily rely on microelectronics that perform complex tasks and rapidly process large data[2]. The functionality, robustness, and longevity of these devices depend on the efficient thermal management of microelectronics as they are becoming increasingly miniaturized and integrated into multi-functional systems [3, 4]. Due to the space limitations inherent in these systems, conventional cooling strategies that rely on heat dissipation via bulky solid materials are impractical. To address the challenges posed by the spatial constraints, single-phase microfluidic cooling techniques are proposed which use liquid coolants to transfer heat away from the source and then dissipate it through a heat exchanger. These techniques provide a number of benefits such as maintaining better temperature uniformity across the heating surface [5] and minimizing pressure drop inside the system [6].

---

[†]Corresponding authors: won@uci.edu, raminb@uci.edu.

While single-phase liquid cooling solutions offer better performance compared to traditional air cooling methods, they entail additional challenges such as requiring high amounts of pumping power to sustain fluid flow through the system. Compared to single-phase cooling solutions, two-phase approaches [7] have the potential to dissipate more heat from electronic devices by leveraging the liquid-vapor phase change processes. The key to enhancing heat transfer performance in two-phase solutions is to utilize structured porous surfaces. Examples of such surfaces include microchannels, micropillars, metal foams, inverse opals, and sintered metal particles which increase thermal efficiency by extending the liquid-solid contact area, supplying capillary-driven flow, accelerating bubble nucleation during boiling, enhancing heat transfer coefficients, or combinations thereof. Aligned with these investigations, [8, 9] focus on incorporating composite porous architectures including micro/nanoparticles and graphene nanoplatelets (GNP)/copper into cooling systems. The empirical findings of these studies report a notable elevation in the heat transfer coefficient (HTC) that ranges from 50% to as high as 290% when compared to systems featuring non-porous and plain surfaces.

The cooling performance of porous microstructures depends on mass and heat transfer which are competing objectives: while pores and their interconnections (characterized by permeability) enable fluid transfer (which, in turn, increases heat dissipation through convection), the solid phase decreases the temperature gradients in the microstructure and is also responsible for most of the me-chanical properties of the structure. To design porous microstructures that strike a balance between these two competing objectives (mass and heat transfer in this case), many prior methods have used task-specific design frameworks that predominantly rely on ad-hoc procedures, see Figure 1. For example, [10] perform a sequence of tests by manually and exhaustively exploring the impact of pillar diameter, height, and center-to-center spacing on the thin-film evaporation performance of a silicon micropillar wick. As another example, [11] apply a similar manual parameterization to explore the pool boiling performance of inverse opal porous structures. These ad-hoc techniques typically yield sub-optimal outcomes due their high costs (associated with developing the ad-hoc framework and collecting data) and their inability to systematically identify and optimize a diverse set of design variables.

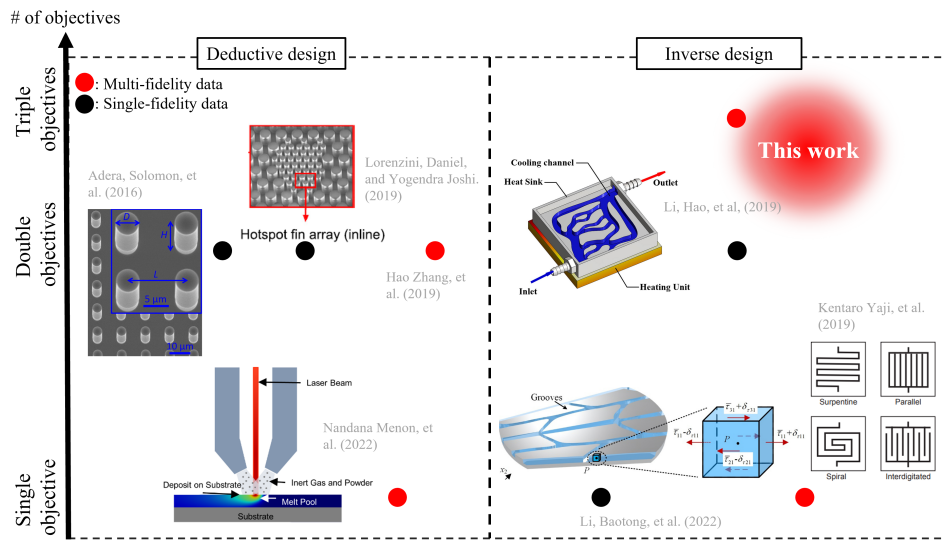

**Figure 1 Schematic comparison between our work and existing methods**: In contrast to current practice, our inverse design framework considers three objectives (permeability, conductivity, and manufacturability) and can leverage an arbitrary number of data sources with any level of accuracy (we use three fidelity levels in this paper).

To address these challenges, we develop a data-driven framework to automate the design of porous microstructures for microelectronic cooling applications. As shown in Figure 2, our framework consists of three primary modules that (1) parameterize the large design space via a handful of interpretable variables, (2) quantify the properties and manufacturing feasibility of design candidates via offline high-fidelity (HF) and low-fidelity (LF) simulations, and (3) build accurate emulators that are used for identifying Pareto optimal designs.

In order to tractably explore a wide variety of microstructure designs, we introduce spectral density functions (SDFs) that represent microstructures in the frequency space. As explained in Section 2.1, we design SDFs that are parameterized via a few physically meaningful variables that can be directly related to microstructural features such as the size and distribution of pore channels [12]. Additionally, microstructures and their SDF representations can be rapidly converted to each other [13], so the computational costs associated with representing the design space via the parameterized SDFs are negligible. These advantages motivate us to search for the optimal design in the parameterized SDF space rather than in the very high-dimensional microstructural space.

Another unique advantage of our framework is its ability to leverage multi-fidelity datasets which dramatically reduces the overall data collection costs during the design process. In most multi-physics problems, there are several available simulation tools which provide different levels of fidelity (i.e., accuracy). This fidelity level primarily depends on the approximations and assumptions that are adopted in the simulations to decrease the computational costs, i.e., the fidelity and costs are highly correlated. Hence, one can accelerate the design process through multi-fidelity modeling [14, 15] where inexpensive LF data are leveraged to reduce the reliance on expensive HF data.

Examples of numerical methods for extracting microstructural thermofluidic properties are pore network modeling (PNM) [16], the finite volume method (FVM) [17], and the lattice Boltzmann method (LBM) [18]. Among these, PNM stands out for its computational efficiency and capability to construct pore networks that faithfully represent the microstructure. In contrast, FVM is one of the most commonly employed techniques for robustly modeling porous structures but it is less efficient than PNM. Finally, LBM is an accurate but computationally very intensive approach. In this work, we employ LBM and PNM to simulate mass and heat transfer, respectively, to strike a balance between efficiency and accuracy. Specifically, we simulate mass transfer in the void portion of porous structures via LBM and simulate heat transfer within the solid portion of the porous structure using PNM. We extract each property set at multiple levels of fidelity by calculating the properties at different resolutions[1], which can be easily controlled in our SDF-based microstructure reconstruction algorithm. Once we build our multi-fidelity dataset, we learn the relation between the SDF parameters and microstructural thermofluidic properties at different resolutions via machine learning (ML) models that are then used to identify Pareto-optimal microstructures and their manufacturing feasibility.

The remainder of the paper is organized as follows. We provide some technical background relevant to our framework in Section 2 and detail our approach in Section 3. We present the results of applying our framework to the problem of porous microstructure design for thermal management in Section 4 and conclude the paper in Section 5.

## 2 Technical Preliminaries

In this section, we first review the fast SDF-based reconstruction method proposed in [13] and then provide a brief overview of latent map Gaussian processes (LMGPs) and their application to multi-fidelity modeling.

### 2.1 Microstructure Generation via Spectral Density Functions

By treating microstructures as realizations of linear shift invariant (LSI) systems, [13] develop a fast reconstruction algorithm to build statistically equivalent microstructures from a target SDF. To demonstrate, we note that the output $Y$ of a given LSI system to any input $X$ is characterized as:

$$S_Y(\nu) = |H(\nu)|^2 S_X(\nu) \tag{1}$$

where $\nu$ is spatial frequency, $S_Y$ and $S_X$ are the SDFs of the input and output signals, and $H$ is the Fourier transform of the impulse response of the system. By treating a microstructure as a discrete signal, specifically the response of an LSI system with white noise input, [13] recast Equation (1) as:

$$\mathscr{S}_R = \mathscr{S}_T \odot \mathscr{S}_W \tag{2}$$

where $\mathscr{S}_R$, $\mathscr{S}_T$, and $\mathscr{S}_W$ denote the reconstructed, target, and white noise discrete SDFs (3-D arrays), respectively, and $\odot$ is the Hadamard product. Given Equation (2) and noting that an SDF is the squared amplitude of a Fourier transform, the microstructure corresponding to $\mathscr{S}_R$ can be reconstructed via:

$$\mathscr{M}_R = \left| \mathscr{F}^{-1}\left\{\sqrt{\mathscr{S}_T}\right\} \odot \mathscr{F}\left\{\mathscr{M}_W\right\} \right| = \mathscr{F}^{-1}\left\{\sqrt{\mathscr{S}_R}\right\} \tag{3}$$

where $\mathscr{M}_R$[2] and $\mathscr{M}_W$ denote the reconstructed and white noise microstructures and $\mathscr{F}\{\cdot\}$ denotes the Fourier transform. In this case, $\sqrt{\mathscr{S}_T} = |\mathscr{F}\{\mathscr{M}_T\}|$ is the transfer function of the system and corresponds to $H$ in Equation (1) which defines the statistical properties of $\mathscr{M}_R$ (i.e., the response of the system to $\mathscr{M}_W$). The significance of Equation (3) is that by changing $\mathscr{M}_W$ one can rapidly generate a large number of microstructures whose stochastic nature is governed by $\mathscr{S}_T$. In this work, we leverage this knowledge and design two parameterized classes of $\mathscr{S}_T$. These parameters are physically meaningful (i.e., they are directly related to the microstructural features) and enable us to systematically build a wide range of microstructures by varying them.

---

[1] By resolution, we mean number of voxels per edge of our cubic microstructure arrays.
[2] Note that $\mathscr{M}_R$ is an array of continuous real values and hence must be level cut to obtain a binary or two-phase microstructure. The threshold used in level cutting is based on the desired phase volume (area in $2D$) fractions.

The above reconstruction method is best suited for quasi-random microstructures whose morphologies are governed by underlying frequency-based correlations. Additionally, reconstructed microstructures may be isotropic or anisotropic depending on the defined target SDF and are periodic (i.e., each face of a microstructure connects seamlessly to its opposite face) which enables easy assembly of a macro-scale structure. However, certain types of microstructures (typically highly ordered ones) cannot be directly modeled via this method due to the use of white noise and the loss of phase information in frequency space during the transformation from Fourier transform to SDF. We refer the reader to [12] for more details on SFD-based microstructure reconstruction.

## 2.2 Latent Map Gaussian Process

LMGPs [14, 19] are extensions of Gaussian processes (GPs) that can handle categorical inputs and which can be applied for multi-fidelity modeling. LMGPs accommodate categorical inputs by learning a parametric function (e.g., a matrix or a feedforward neural network) that maps a prior representation of categorical variables into a low-dimensional quantitative embedding (latent space) which, in turn, allows us to directly use the categorical variables in a GP.

We denote the mapping function of an LMGP[3] via $\boldsymbol{\zeta}(\boldsymbol{\iota}) : \mathbb{R}^{d\iota} \to \mathbb{R}^{d\zeta}$ where $d\zeta$ is the dimensionality of the latent space and $\boldsymbol{\iota} = [\iota_1, \ldots, \iota_{d\iota}]^T$ are the appropriately encoded (e.g., one-hot) categorical inputs of dimensionality $d\iota$. The correlation between two inputs $(\boldsymbol{\xi}, \boldsymbol{\iota})$ and $\left(\boldsymbol{\xi}', \boldsymbol{\iota}'\right)$ can then be found by modifying any standard GP correlation function $r(\cdot, \cdot)$ to accept the (now quantitative) mapped categorical variables. A common choice of correlation function is the Gaussian, which is modified as:

$$r\left((\boldsymbol{\xi}, \boldsymbol{\iota}), (\boldsymbol{\xi}', \boldsymbol{\iota}')\right) = \exp\left(-\sum_{i=1}^{d\xi} 10^{\omega_i}(\xi_i - \xi_i')^2\right) \times \exp\left(-\left\|\boldsymbol{\zeta}(\boldsymbol{\iota}) - \boldsymbol{\zeta}(\boldsymbol{\iota}')\right\|_2^2\right) \quad (4)$$

where $\boldsymbol{\xi} = [\xi_1, \xi_2, \ldots, \xi_{d\xi}]^T \in \mathbb{R}^{d\xi}$ are the numerical inputs, $\|\cdot\|_2$ denotes the Euclidean 2-norm, and $\boldsymbol{\omega} = [\omega_1, \ldots, \omega_{dx}]^T$ are the roughness or scale parameters.

As indicated by Equation (4), distances in the latent space directly correspond to correlation between categorical inputs. Specifically, if $\boldsymbol{\zeta}(\boldsymbol{\iota})$ and $\boldsymbol{\zeta}(\boldsymbol{\iota}')$ are mapped to two latent points whose distance is $\Delta$, then LMGP indicates that the correlation between these two categorical combinations is $\exp\left(-\Delta^2\right)$. That is, plotting $\boldsymbol{\zeta}(\boldsymbol{\iota})$ for every categorical combination after fitting an LMGP provides a visual interpretation of the relationships between the categorical variables and their levels. We use this visually interpretable result of LMGP in Section 4.1 to validate our emulators' ability to correctly learn the relationships within our dataset.

Multi-fidelity modeling via LMGP starts with augmenting the data sets with a categorical variable whose levels denote the data sources. Then, all the data sets are combined and an LMGP is fitted directly to the unified data set. As described above, this method provides a visual representation of the relationships between the data sources. As detailed in [14], multi-fidelity modeling via LMGPs provides a number of unique advantages such as (1) the ability to jointly fuse data from an arbitrary number of sources which may have categorical features, (2) accurate and probabilistic predictions for all data sources, (3) interpretability via the latent space and other model hyperparameters post-fitting, (4) the ability to incorporate prior knowledge through choice of kernel / correlation function, and (5) insensitivity of the model's performance to the majority of the tuning parameters. We refer the interested reader to [14, 15, 20] for more details on multi-fidelity modeling via manifold learning approaches such as LMGP.

---

[3]LMGP also accepts using multiple mapping functions to individually map subsets of the categorical combinations, e.g., we use two mapping functions for some of our emulators in Section 3.3.

## 3 Methodology

Direct optimization of high-resolution microstructures (e.g., via topology optimization) that provide desired property sets is quite challenging as property evaluations are complex and computationally costly. To address this challenge, we develop an indirect inverse design framework where the optimization is performed in a reduced dimensional space, see Figure 2. As detailed in Section 3.1, we build this low-dimensional design space by encoding a microstructure (which can be isotropic or anisotropic) via a few parameters that characterize its corresponding SDF. We also develop objective functions that quantify manufacturability in addition to directional properties (permeability and thermal conductivity, see Section 3.2.3) of a microstructure. To connect these objectives to the design variable (i.e., the SDF parameters), we leverage multi-fidelity emulation where low-fidelity data (based on coarse microstructure discretization) are used to reduce the reliance on expensive high-fidelity samples. Once the emulators are built, we identify the Pareto-optimal microstructures via the Genetic Algorithm (GA) as detailed in Section 3.3.

### 3.1 Design Space Encoding

To tractably explore the microstructure design space we dramatically reduce its dimensionality via SDFs because (1) conversion between a microstructure and its SDF representation is computationally very fast (see Section 2.1), and (2) SDFs are relatively well ordered for quasi-random microstructures and hence can be readily parameterized with a few variables [12, 13] which can be easily linked to morphological features such as degree of anisotropy/isotropy, size and distribution of pore networks, or anisotropy direction, see Figure 4.

The SDF representation of a microstructure is a 3-D array[4] whose nonzero elements determine its shape and the corresponding microstructure morphology. SDFs that are used for designing microstructures are often generated as a binary array (all elements are only $0$ or $1$) for simplicity, but we may represent a greater variety of microstructure morphologies by defining the magnitude of the nonzero elements of the SDF as a function of radius (distance from the origin) via the 1-D power spectrum [12]. Power spectrum shapes can be modeled via, e.g., triangle/flat waves[5] or the delta function.

There are many classes of SDF shapes with associated parameterizations, e.g., spheres in 3-D which may be defined by just the radius. Each choice of combination of SDF shape and 1-D power spectrum may represent some restricted subset of possible microstructure designs. Here, the challenge is to choose a combination that reduces the dimensionality of the design space while maintaining sufficient representation power to cover the space of optimal designs for the target application. To address this challenge, we develop two types of SDF shapes which extend 2-D partial rings[6] to portions of a sphere (denoted by *Sph*) or portions of a cylinder (denoted by *Cyl*). To construct an SDF via either parameterization, we begin by forming a 2-D partial ring in the $x$-$y$ plane by defining a point at radius $r$ on the $x$ axis, sweeping through an angle $\theta$ in both the positive and negative directions, and reflecting the resulting arc across the $y$ axis.[7] The arcs are then either rotated about the $y$ axis by an angle $\phi$ in both the positive and negative directions to form a *Sph*-type SDF, as shown in Figure 3a, or are extruded vertically in both the positive and negative directions by a height $h$ to form a *Cyl*-type SDF, as shown in Figure 3b.

---

[4]In this paper, we assume that the frequency space has been shifted to be zero centered, i.e., the center of the array corresponds to zero frequency, and we consider the center of the array to be the origin in our coordinate system.

[5]Using a flat wave 1-D power spectrum is equivalent to a using binary SDF.

For the 1-D power spectrum, we use the Gaussian probability density function (PDF),

$$f(\varsigma) = \frac{1}{\sigma\sqrt{2\pi}} \exp\left\{-\frac{1}{2}\left(\frac{\varsigma - r}{\sigma}\right)^2\right\} \tag{5}$$

---

[6] A shape commonly seen in 2-D slices of real microstructures.

[7] The reflection across the $y$ axis, while not strictly necessary, makes user-designed microstructures more closely resemble those derived from real microstructures, which have discrete SDFs (via fast Fourier transform) that are symmetric with respect to radius in spherical coordinates.

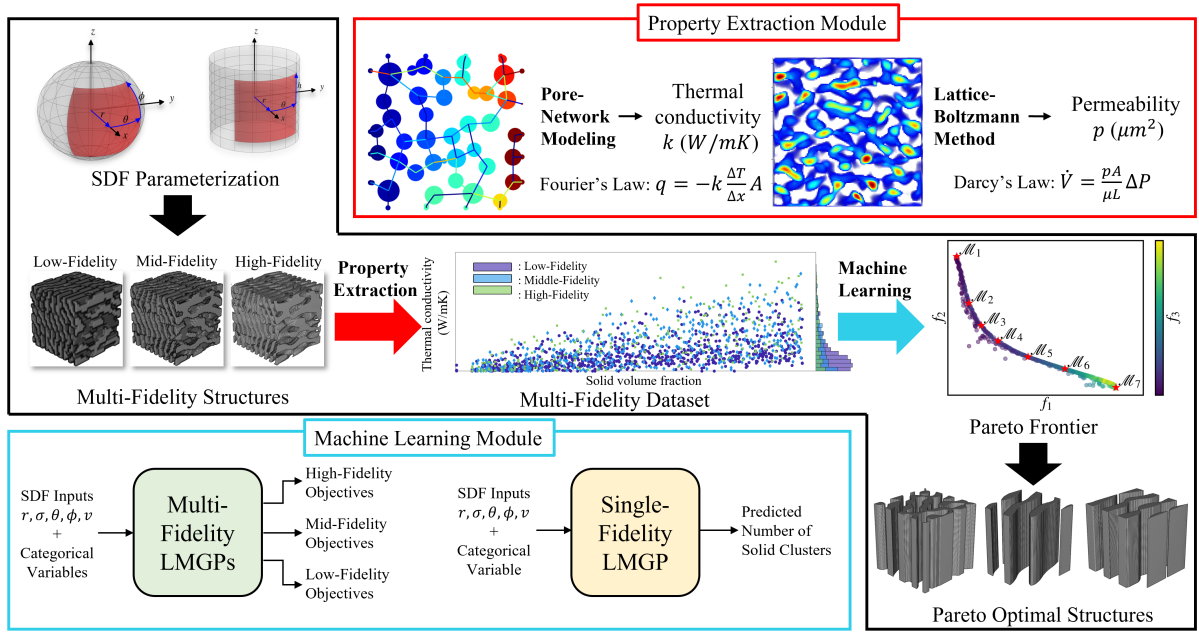

**Figure 2 Multi-fidelity multi-objective framework for the design of porous microstructures:** We automate the design of porous microstructures via three main modules. First, we develop two SDF parameterizations and use them to tractably explore the design space at three fidelity levels. Then, we extract thermofluidic properties and feasibility information via simulations and develop objectives as functions of these properties. Finally, identify Pareto-optimal microstructure designs by building emulators that predict the properties and manufacturing feasibility of the reconstructions.

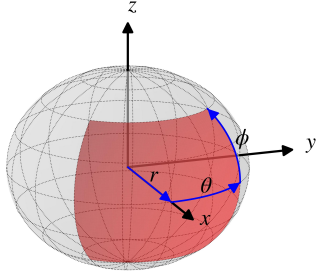
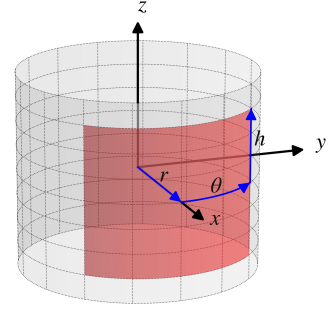

**(a)** Sphere-type SDF parameterization   **(b)** Cylinder-type SDF parameterization

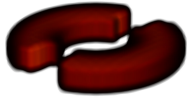
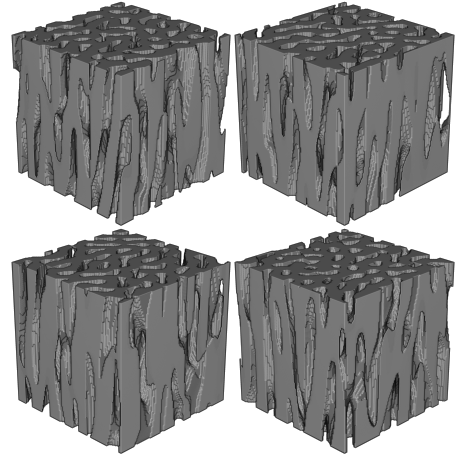

**(c)** Example of cylinder-type SDF   **(d)** Microstructure realizations from example SDF

**Figure 3 Microstructure reconstruction via parameterized SDFs: (a)** A point at radius $r$ on the $x$ axis is swept through $\pm\theta$ about the $z$ axis to create an arc which is then swept through $\pm\phi$ about the $y$ axis to create a curved rectangular patch with two axes of curvature. **(b)** A point at radius r on the $x$ axis is swept through $\pm\theta$ about the $z$ axis to create an arc which is then extruded vertically by a height $h$ in the $\pm z$ direction to create a curved rectangular patch with one axis of curvature. For clarity, we do not show the final reflection with respect to the radius (across the $y - z$ plane) in **(a)** and **(b)**. **(c)** An example of a squat cylinder-type SDF produced from our parameterization. **(d)** Four microstructure realizations from the squat cylinder-type SDF demonstrating anisotropic behavior.

where $\varsigma$ is the distance from the origin[8], $r$ is the chosen radius[9], and $\sigma$ is the standard deviation. The Gaussian PDF can roughly approximate a triangle wave, flat wave, or delta function with appropriate choices of $r$ and $\sigma$. Our choice of the 1-D power spectrum combined with our two parameterized SDF shapes provide high representation power and allow us to explore a wide variety of microstructure designs.

Once we choose some values for the SDF parameters (e.g., $r = 7.40, \sigma = 1.22, \theta = 1.25, \phi = 0.30$ for a *Cyl* type SDF) and build the SDF as demonstrated in Figure 3, the corresponding mi-crostructure is realized with the desired volume fraction $v$[10] as described in Section 2.1. We show some examples of the types of SDFs and corresponding structures generated via our parameterizations in Figure 4. As illustrated, we generate a wide range of quasi-random structures with various degrees of anisotropy and order based on the choice of parameters. Note that we hold $v$ constant at $0.5$ in Figure 4 in order to highlight the effect of the choice of SDF—we can generate an even wider range of behaviors with varying the values of volume fraction (which we do in Section 4).

---

[8] Or the distance from the $z$ axis for the cylinder-type SDF.
[9] The same radius used in defining the SDF shape.

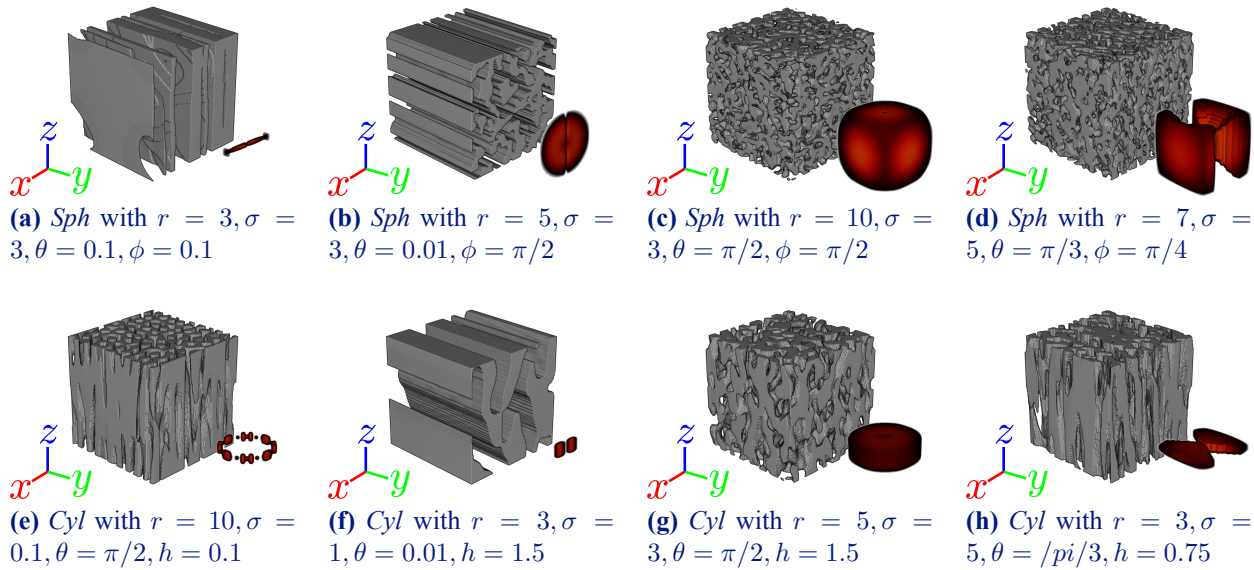

(a) *Sph* with $r = 3, \sigma = 3, \theta = 0.1, \phi = 0.1$.

(b) *Sph* with $r = 5, \sigma = 3, \theta = 0.01, \phi = \pi/2$.

(c) *Sph* with $r = 10, \sigma = 3, \theta = \pi/2, \phi = \pi/2$.

(d) *Sph* with $r = 7, \sigma = 5, \theta = \pi/3, \phi = \pi/4$.

(e) *Cyl* with $r = 10, \sigma = 0.1, \theta = \pi/2, h = 0.1$.

(f) *Cyl* with $r = 3, \sigma = 1, \theta = 0.01, h = 1.5$.

(g) *Cyl* with $r = 5, \sigma = 3, \theta = \pi/2, h = 1.5$.

(h) *Cyl* with $r = 3, \sigma = 5, \theta = /pi/3, h = 0.75$.

**Figure 4 Examples of microstructures generated from our SDF parameterization:** We can generate a wide variety of microstructures using our two SDF parameterizations despite only using four parameters $r, \sigma, \theta,$ and $\phi$ (or $h$) to characterize the structure ($v$ is held constant at $0.5$). Microstructures are shown with their corresponding SDF (*Sph*-type for **(a)**-**(d)** and *Cyl*-type for **(e)**-**(h)**) in the bottom-right corner.

Our proposed SDF parameterizations each reduce the input space to a tractable dimensionality—both require only five input parameters to define an SDF and its reconstructed microstructure, specifically $r, \sigma, \theta, \phi$ (or $h$), and $v$. While it is straightforward to separately explore each individual parameterization efficiently via the design of experiments (DOE), we aim to explore both at once to take full advantage of the space-filling properties of quasi-random sequences. To accomplish this, we unify the two input spaces via the transformation $h = r\tan(\phi/2)$ and add an additional input $\mathcal{T}$ which denotes the SDF type (*Sph* or *Cyl*). We then explore the unified input space via 6-dimensional Sobol sequence, rounding the input dimension corresponding to the SDF-type to convert to a categorical input. We refer to this collection of 6 parameters which defines our parameterized SDF as the design variables.

## 3.2 Design Evaluation

The thermofluidic system of interest in this work is a microelectronic device cooled by an evaporator wick subjected to a uniform heat flux from its bottom. The porous microstructure in the evaporator wick, which can be either isotropic or anisotropic, remains saturated and enveloped in water, see Figure 5. The influx of liquid into the porous microstructure is passively driven by capillary forces and leads to evaporation at the top where the to-be-dissipated heat is removed through a phase change process. Modeling this system relies on considering intricate interactions between mass transfer, heat transfer, and phase change within the porous microstructure composing the wick and is therefore quite challenging. To tackle this problem, we begin by making the well-motivated (see Section 3.2.2) assumption that all of the fluid that is passed through the porous structure completely evaporates at its top surface (this assumptions allows us to establishe a clear boundary condition). With this premise, cooling becomes primarily a function of in-plane mass transfer and out-of-plane heat transfer. Therefore, we quantify directional permeabilities and thermal conductivities as the key descriptors for evaluating the overall thermofluoidic performance of the evaporator wick.

---

[10]In this paper, we consider only two-phase microstructures consisting of solid copper and void.

As detailed below, we extract these directional permeabilities and thermal conductivities via LBM and PNM, respectively. In these simulations, we employ the microstructures that are built via the process described in Section 3.1. Additionally, we note that in this section and the following ones we distinguish between the coordinate system used for the SDF/microstructure $(x, y, z)$ and the one used for the thermofluidic system $(x^{'}, y^{'}, z^{'})$. We adopt this distinction because we can align a generated microstructure in three different directions within the evaporative wick system.

### 3.2.1 Permeability Modeling

In this section, we develop a customized model based on LBM to simulate the mass transfer. LBM is well suited for simulating flow-based phenomena in complex geometries such as porous microstructures. It is based on the concept of simulating the fluid as a large number of particles that move and collide with each other over a discrete lattice mesh. In particular, the Boltzmann equation describes the statistical behavior of the dynamic particle system where particle distributions are utilized to compute the fluid velocity and pressure profiles [21]. We apply a single-time relaxation scheme based on the Bhatnagar-Gross-Krook (BGK) collision operator to calculate the permeability [22, 23]. We use the standard BGK version of this operator with a D3Q19 lattice which requires the particles that represent the fluid flow to satisfy the distribution equation $f(x^{'}, t)$ formulated as:

$$f(x^{'} + e_i, t) = \frac{1}{\tau}(f(x^{'}, t) - f_{eq}(x^{'}, t) = \Omega(x^{'}, g, \tau, t) \quad (6)$$

where $e_i$ is the particle velocity in the $i^{th}$ direction, $t$ is the simulation time, $\tau$ is the relaxation time, and $\Omega$ is the collision operator.

While simulating the mass transfer we employ Dirichlet boundary conditions by defining constant pressure drops across opposing faces and keeping all other boundary faces impermeable. Since our design space includes anisotropic microstructures, we implement six simulations for each microstructure by rotating it about different axes. The outcomes of the six simulations provide three pairs of identical permeability values which indicates that swapping the inlet and outlet faces has no influence on the steady-state directional permeability calculations[11]. Based on this expected finding, henceforth we only run three simulations for a microstructure.

In each of our mass transfer simulations we set the constant pressure drop equal to the capillary pressure $P_{cap}$ which varies based on the topology of different porous microstructures. Assuming a

---
[11]Since these simulations are iterative, the equality of these permeability values is checked upon convergence.

spherical meniscus, we use the Young-Laplace equation to express $P_{cap}$ as:

$$P_{cap} = \frac{2\sigma' cos(\theta)}{R} \quad (7)$$

where $\sigma'$ is the water surface tension in $N/m$, $\theta$ is the contact angle in degrees, and $R$ is the average pore radius of the porous medium [24]. Once the velocity field is extracted from LBM, we calculate the directional permeabilities via Darcy's law:

$$\dot{V} = \frac{p}{\mu L} \triangle P A \quad (8)$$

where $\dot{V}$ is the fluid volumetric flow rate in $m^3/s$, $p$ is the permeability in $m^2$, $A$ is the cross-sectional area of the fluid flow in $m^2$, $\mu$ is the fluid dynamic viscosity in $Pa \cdot s$, $L$ is the length of the medium in $m$, and $\triangle P$ is the inlet to outlet pressure drop in $Pa$.

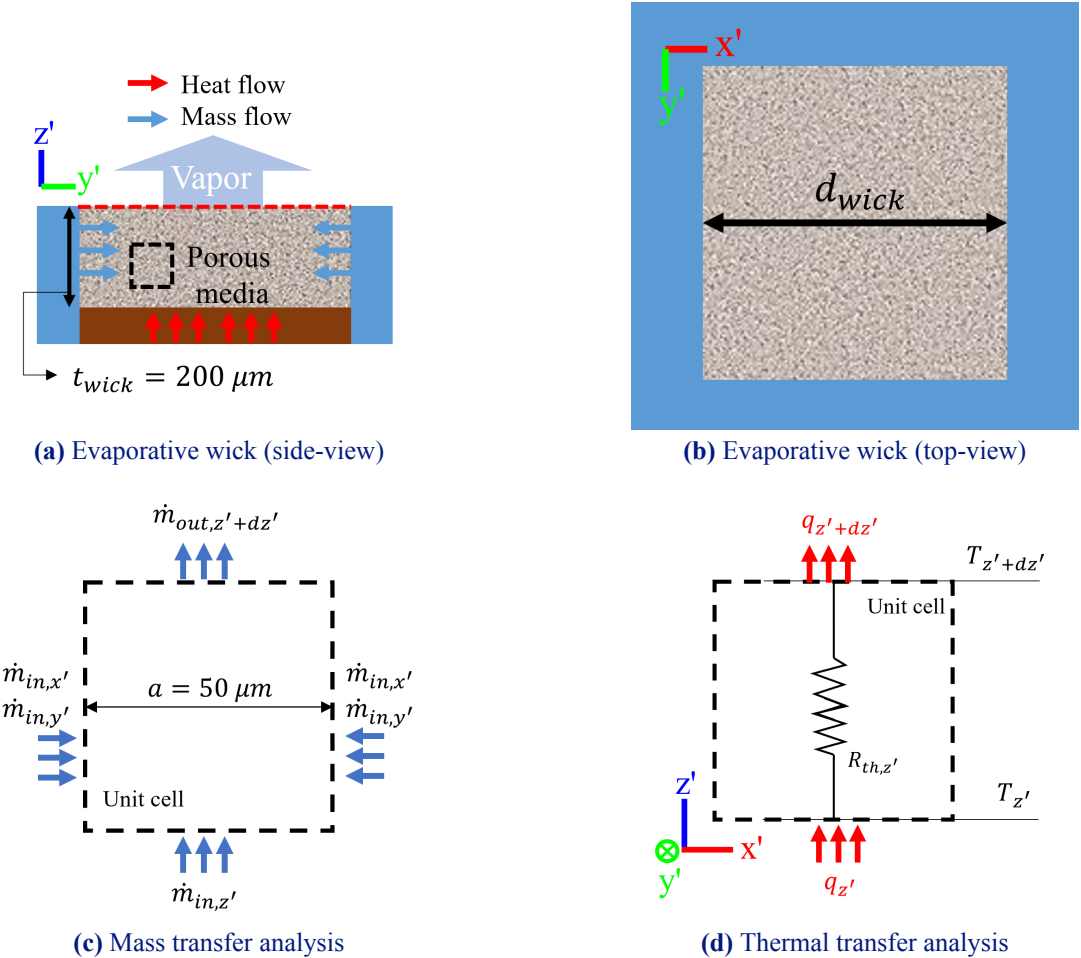

**Figure 5 Mass and heat transfer within an evaporative wick: (a)** and **(b)** The water pool surrounds the wick and the to-be-dissipated heat is generated at the bottom of the wick. The black dashed box in **(a)** represents the microstructure that tessellates the porous medium and the red dashed line in **(a)** is the evaporating surface of the wick. **(c)** and **(d)** represent, respectively, the mass and thermal analysis of the region specified with the black-dashed line in **(a)** that has a side length of $a = 50~\mu m$.

### 3.2.2 Thermal Conductivity Modeling

We model thermal conductivity via PNM which is frequently used to simulate mass and heat transfer within porous microstructures such as rocks [25], soil, and membranes [26]. PNM creates a simplified network representation of the real pore space where the pores and their connections are represented by, respectively, spheres and sticks. We use the open-source software PoreSpy [27] to build the pore network of the microstructures. Specifically, we first transfer the reconstructed microstructures to boolean-type numpy arrays which are then processed via the sub-network of an over-segmented watershed (SNOW) algorithm (for noise reduction, thresholding, and segmentation) to distinguish the pore space from the solid phase. Following this processing step, spherical pores and cylindrical throats (the connections between pores) are identified and combined into an interconnected PNM, which provides a simplified representation of the porous microstructure. We note that PNMs model mass and heat transfer in pores and hence we model the heat transfer in the solid portion of the porous microstructure by swapping the phases

We use OpenPNM [28] to model heat transfer. We first categorize the six faces of a microstructure with descriptive labels: *left* and *right* (corresponding to the inlet and outlet along the $x'$ direction), *front* and *back* (along the $y'$ direction), and *top* and *bottom* (along the $z'$ direction). Then, we impose Dirichlet boundary conditions on the opposing faces, define a constant temperature drop of $\Delta T = 50K$, and maintain the remaining boundary faces in thermally insulated states, see Figure 5. After imposing the boundary conditions, we discretize the PNM into a series of small control volumes that comprise of two pores and their connecting throat. We then employ the Fourier conduction law to solve the energy balance equation in the control volume given PNM-based properties including average temperatures of two pores, throat cross-section area, and heat flow rate in the throat:

$$q = -k \frac{\partial T}{\partial n} A \tag{9}$$

where $q$ is the heat rate in $W$, $n$ is the spatial variable including $x'$, $y'$, and $z'$ directions, $T$ is the temperature in $K$, and $k$ represents the effective thermal conductivity of the microstructure in $W/mK$. Finally, the thermal conductivity in a specific direction of the microstructure is obtained by solving a system of linear energy balance equations at each control volume in the PNM.

### 3.2.3 Design Objectives

We develop two objective functions to evaluate the thermofluidic performance of a microstructure and an additional one to estimate its feasibility which we define as the absence of floating solid clusters disconnected from the boundaries. The first objective, $f_1$, measures the heat dissipation capacity based on the mass transfer of liquid to the top surface of the microstructure. In this study, we make the simplifying assumption that all the water transported from the surrounding water pool through the porous microstructure—driven by capillary forces—eventually evaporates at its top surface. This assumption is validated by the comparatively elevated heat dissipation rates associated with thin-film evaporation on the top surface of the porous structure. The heat rates based on phase change mechanism are orders of magnitude (ranging from 10 to 100 times) higher than those facilitated solely by mass transfer of liquid through the porous structure. Hence, $f_1$ quantifies the ability of the microstructure to transport fluids, which determines the thin-film evaporation rate at the top surface of the porous structure.

The second objective, $f_2$, measures the microstructure's capability to maintain minimal temperature gradients in it. Given the previously established dominance of heat transfer in $z'$ direction, it is logical to focus on the thermal properties in this particular direction where larger temperature gradients are anticipated. As a result, we define $f_2$ as the reciprocal of thermal resistance in the out-of-plane ($z'$) direction of the microstructure and derive both objectives below.

The volumetric flow rate through a porous microstructure can be calculated using Darcy's law in Equation (8). As shown in Figures 5a and 5c, the fluid enters the microstructure (represented via the black-dashed unit cell) from both $x'$ and $y'$ directions. The mass flow rates $\dot{m}_{in,x'}$ and $\dot{m}_{in,x'}$ in $x'$ and $y'$ directions are:

$$\dot{m}_{in,x'} = \rho \dot{V}_{in,x'} \tag{10}$$

$$\dot{m}_{in,y'} = \rho \dot{V}_{in,y'} \tag{11}$$

where $\dot{m}_{in,x'}$ and $\dot{m}_{in,y'}$ are the mass flow rates in $kg/s$ and $\rho$ is the liquid density in $kg/m^2$. To avoid wick dry-out, we set the pressure drop $\triangle P_{tot}$ equal to the capillary pressure $P_{cap}$ to balance the supplied and evaporated liquid through the whole wick, that is:

$$\triangle P_{tot} = P_{cap} \tag{12}$$

Combining Equations (7) and (12), the pressure drop in the microstructure can be expressed as:

$$\triangle P_{uc} = \frac{2\sigma \cos(\theta)}{R} \frac{a}{d_{wick}} \tag{13}$$

where $\triangle P_{uc}$ is the pressure drop through the unit cell, $R$ is the average pore size, $a$ is the side length of the cubic unit cell, and $d_{wick}$ is the length of the wick. Combining Equation (8) and Equations (10) - (13), we can express $\dot{m}_{in,x'}$ and $\dot{m}_{in,x'}$ as:

$$\dot{m}_{in,x'} = \rho \frac{p_{x'}}{\mu/2} \triangle P_{uc} a \tag{14}$$

$$\dot{m}_{in,y'} = \rho \frac{p_{y'}}{\mu/2} \triangle P_{uc} a \tag{15}$$

Using the conservation of mass, $\dot{m}_{in} = \dot{m}_{out}$, we can write:

$$2\dot{m}_{in,x'} + 2\dot{m}_{in,y'} + \dot{m}_{in,z'} = \dot{m}_{out,z'+dz'} \tag{16}$$

To obtain the mass flow rate difference in the $z'$ direction $\triangle \dot{m}_{z'}$, we rearrange Equation (16) as:

$$\triangle \dot{m}_{z'} = \dot{m}_{out,z'+dz'} - \dot{m}_{in,z'} = 2\dot{m}_{in,x'} + 2\dot{m}_{in,y'} \tag{17}$$

and express $\triangle \dot{m}_{z'}$ as:

$$\triangle \dot{m}_{z'} = 4\rho \frac{p_{x'} + p_{y'}}{\mu} \triangle P_{uc} a. \tag{18}$$

As shown in Figure 5d, the heat flow into and out of the microstructure are $q_{z'}$ and $q_{z'+dz'}$ in $W$, respectively. We apply Fourier's conduction law in this direction as shown below:

$$q_{z'} = -k_{z'} \frac{dT}{dz'}\Big|_{z'} a^2 \tag{19}$$

$$q_{z'+dz'} = -k_{z'}\frac{dT}{dz'}\Big|_{z'+dz'}a^2 \tag{20}$$

where $k_{z'}$ is the $z'$-direction effective thermal conductivity of the microstructure in $W/mK$. Hence, the dissipated heat in the $z'$ direction is:

$$\triangle q_{z'} = q_{z'+dz'} - q_{z'} = -k_{z'}\frac{T_{z'+dz'}-T_{z'}}{a}a^2 \tag{21}$$

based on thermal circuit analysis, which can also be expressed as:

$$\triangle q_{z'} = \frac{T_{z'} - T_{z'+dz'}}{R_{th,z'}} \tag{22}$$

where $R_{th,z'}$ is the z-direction thermal resistance in K/W. By combining Equations (21) and (22), we can write $R_{th,z'}$ as:

$$R_{th,z'} = \frac{1}{k_{z'}a} \tag{23}$$

Since we assume that all the mass flow coming into the wick evaporates at its top surface, $\triangle \dot{m}_{z'}$ must be maximized. Additionally, to minimize the temperature gradient across the wick in the out-of-plane direction, the effective thermal resistance in the z-direction should be minimized. Hence, our two objectives $f_1$ and $f_2$ are defined as:

$$f_1 = 4\rho\frac{p_{x'} + p_{y'}}{\mu}\triangle P_{uc}a \tag{24}$$

$$f_2 = k_{z'}a \tag{25}$$

The two objectives in Equation (24) and Equation (25) are defined in terms of the thermofluidic system's coordinates or $x'$, $y'$, and $z'$ where $x'$ and $y'$ are treated identically due to symmetry. Hence, a given microstructure can be placed in the system in one of three possible non-degenerate orientations $\mathcal{O}_1$, $\mathcal{O}_2$, or $\mathcal{O}_3$ corresponding, respectively, to placing the microstructure such that its $z$, $y$, or $x$ axis is aligned with the $z'$ axis of the system. Note that, for SDFs (and hence microstructures) generated via the *Sph* parameterization, $\mathcal{O}_1$ is equivalent to $\mathcal{O}_2$ with $\theta$ and $\phi$ swapped and we therefore expect to see identical objective space coverage from these combinations of SDF type and orientation[12]. We test this expectation in Section 4.

As mentioned earlier, we additionally define a third objective, $f_3$, as the expected number of distinct solid clusters in microstructure realizations. We wish to minimize $f_3$ to increase the probability that our designs are physically feasible. As explained in Section 3.3, we estimate $f_3$ of a candidate microstructure's SDF parameterization via emulation. That is, we first leverage our fast reconstruction algorithm to generate a large number of microstructures and then for each sample calculate the expected number of floating solid clusters (which should ideally equal 0). Once this dataset is built, we train a GP that predicts the expected number of solid clusters in a microstructure given its SDF parameterization.

---

[12]We expect distinct behaviors for all three orientations of microstructures generated via the *Cyl* SDF parameterization as there are no symmetries between axes.

### 3.3 Multi-fidelity Emulation and Inverse Design

This study aims to find microstructures which maximize $f_1$ and $f_2$ while minimizing $f_3$. These designs are Pareto optimal in the three-dimensional objective space defined as:

$$\mathcal{F}: \ \boldsymbol{u} \equiv [\boldsymbol{\xi}, \boldsymbol{\iota}] \longmapsto [f_1, f_2, -f_3] \in \mathbb{R}^3 \tag{26}$$

if there exist no choices of $\boldsymbol{u}$ which increase the value of any dimension of $\mathcal{F}$ without decreasing the value of at least one of the other dimensions. The input space $\boldsymbol{u} = [\boldsymbol{\xi}, \boldsymbol{\iota}]$ is 8-dimensional with $\boldsymbol{\xi} = [r, \sigma, \theta, \phi, v, \mathscr{T}]$ as quantitative inputs, where $\mathscr{T}$ is the choice of SDF parameterization which we treat as a numeric variable rather than categorical by assigning $0$ to *Sph* and $1$ to *Cyl*. The remaining two inputs $\boldsymbol{\iota}$ are categorical and correspond to the fidelity level (resolution) and the choice of orientation used when calculating $f_1$ and $f_2$.[13]

Since directly evaluating our objectives via simulations is too expensive, we build emulators $\eta_1(\boldsymbol{u})$, $\eta_2(\boldsymbol{u})$, and $\eta_3(\boldsymbol{\xi})$ to cheaply predict, respectively, $f_1$, $f_2$, and $f_3$, and work in an approximation of our objective space:

$$\hat{\mathcal{F}}: \ \boldsymbol{u} \equiv [\boldsymbol{\xi}, \boldsymbol{\iota}] \longmapsto \left[\hat{f}_1, \hat{f}_2, -\hat{f}_3\right] \equiv [\eta_1(\boldsymbol{u}), \eta_2(\boldsymbol{u}), -\eta_3(\boldsymbol{\xi})] \in \mathbb{R}^3 \tag{27}$$

While emulation reduces the computational costs of our inverse design problem, training accurate models requires large datasets which are still too expensive to construct at a high level of fidelity for $f_1$ and $f_2$. As such, we leverage multi-fidelity modeling to reduce the cost of data generation while achieving the desired degree of accuracy in learning the high-fidelity system. We generate high-, mid-, and low-fidelity microstructures as cubes with edge lengths $125$, $100$, and $75$ voxels, respectively, and build the datasets to train $\eta_1$ and $\eta_2$ by calculating the outputs $f_1$ and $f_2$ as described in Section 3.2.3. Property evaluation is much faster for the lower-fidelity structures, but is correspondingly less accurate. We then train multi-fidelity LMGPs to serve as surrogates for $f_1$ and $f_2$.

Depending on the choice of volume fraction and other input parameters, our generated microstructures may have floating solid clusters that are not connected to any of the edges of the cell. Such microstructures are not manufacturable or physically feasible. To avoid such microstructures in the design process, we process training datasets of $\eta_1$ and $\eta_2$ as follows: for each point in the input space, we generate $50$ microstructure realizations[14] then, if there is no realization without floating clusters, select the microstructure with the smallest volume of floating clusters and remove the small disconnected clusters. If the change in volume fraction due to removing the clusters is very small (e.g., below $5\%$), then we accept the processed structure and add the datum to the dataset (we do change $v$ in the stored input to match its updated value after removing the clusters). If none of the $50$ microstructure realizations result into fully connected designs after the above procedure, we discard the input point.

Unlike $f_1$ and $f_2$, it is relatively cheap to evaluate $f_3$ (although still too expensive to do within a loop). Therefore, we use a single-fidelity LMGP to emulate $f_3$ trained on a high-fidelity dataset of structures with edge length 150. $\eta_3(\boldsymbol{\xi})$ is trained with $\boldsymbol{\xi}$ as its inputs and $f_3$ estimated as the average (over 50 realizations) number of solid clusters as its output. We use only one resolution and orientation has no influence on the number of solid clusters, so we eliminate these categorical variables and train an LMGP using $\mathscr{T}$ as our only categorical variable. We provide more detail on $\eta_1$, $\eta_2$, and $\eta_3$ in Section 4.1.

---

[13] We choose this treatment of the qualitative inputs as it yields the highest cross-validation emulation accuracy, see Section 4.1 for remaining details.

[14] This is computationally tractable because microstructure generation via SDF is many orders of magnitude faster than property extraction.

After constructing our emulators, we find the Pareto-optimal microstructures via GA (we use the Python package pymoo, see Table 1 for the specific GA parameters used). We define the GA search space to be the five quantitative inputs $[r, \sigma, \theta, \phi,$ and $v]$ and fix the other (categorical) variables to search a specific combination of SDF type and orientation.[15] We restrict this search space to be within the range seen in the training data for each quantitative input dimension as GPs (and by extension LMGPs) are inaccurate in extrapolation. Using the GA, we find 500 Pareto-optimal points (we refer to the set of Pareto-optimal points as the Pareto frontier) for each combination of the 3 orientations and 2 SDF types for a total of 3,000 structures. We examine this Pareto frontier and some selected Pareto-optimal microstructures in Section 4.2.

## 4 Results

In this section, we first validate the accuracy of our emulators for each objective and present the results of the multi-objective optimization via GA in Sections 4.1 and 4.2, respectively. In section Section 4.2 we use the NSGA2 genetic algorithm implemented in pymoo for our multi-objective optimization with the parameters shown in Table 1. Then, in Section 4.3, we examine some selected microstructure realizations on the Pareto frontier and compare the LMGP prediction of our objective function values to those predicted from thermofluidic property simulations.

**Table 1** Genetic algorithm parameters used in multi-objective optimization.

| Parameter | Value |
|---|---|
| Population size | 500 |
| Number of offspring | 100 |
| Crossover (SBX) probability | 0.9 |
| Crossover (SBX) eta | 15 |
| Mutation (PM) eta | 20 |
| Number of Generations | 256 |

### 4.1 Emulation Performance

To train $\eta_1$ and $\eta_2$, we generate a multi-fidelity microstructure dataset of size $n_h = 69$, $n_m = 131$, and $n_l = 206$ where $n_h$, $n_m$, and $n_l$ refer to the number of high-, mid-, and low-fidelity data, respectively. To explore a sufficiently large design space, we set the ranges of the variables as $r \in [3, 10]$, $\sigma \in [0.06, 3.0]$, $\theta \in [0.15, \pi/2]$, $\phi \in [0.15, \pi/2]$, and $v \in [0.15, 0.7]$.[16] We then extract properties and calculate $f_1$ and $f_2$ for the three possible orientations, leaving us with a training dataset of size $n_h = 3 \times 69 = 207$, $n_m = 3 \times 131 = 393$, $n_l = 3 \times 206 = 618$. Using a convergence study based on incrementally increasing the portion of the data used in training, we fit $\eta_1$ to a subset of the data with $n_h = 207$, $n_m = 303$, $n_l = 390$ and $\eta_2$ to all data.

---

[15] We fix the categorical variable for fidelity level such that we always predict at the highest fidelity level.

[16] We place nonzero lower bounds on the sample range for $\theta$ and $\phi$ since the rays swept out by the SDF shape at very small values are all "between" the voxel locations in the discrete representation of the SDF, resulting in the absence of nonzero voxels.

Since we use all available high-fidelity data to train $\eta_1$ and $\eta_2$ and do not reserve test data for validation, we rely on 5-fold cross-validation (CV) to assess the accuracy of our emulators. Specifically, we split the training data randomly into 5 partitions and train 5 emulators with each using only $4/5$ of the available data (one for each possible permutation of the 5 folds). We then evaluate accuracy by obtaining predictions for each fold on the remaining $1/5$ of the data and finally calculate the normalized root mean squared error (NRMSE) across all folds as:

$$\text{NRMSE}\left(\boldsymbol{f}_{test}, \boldsymbol{f}_{pred}\right) = \sqrt{\frac{(\boldsymbol{f}_{pred} - \boldsymbol{f}_{test})^T (\boldsymbol{f}_{pred} - \boldsymbol{f}_{test})}{n \times \text{var}(\boldsymbol{f}_{test})}} \qquad (28)$$

where $\boldsymbol{f}_{test}$ is the combined vector of test outputs for all folds, $\boldsymbol{f}_{pred}$ is the combined vector of predicted outputs from the test inputs for all folds, $n$ is the total number of test data, and $\text{var}(\cdot)$ is the variance. We obtain NRMSEs of $0.047$ and $0.089$ for the emulators for $f_1$ and $f_2$, respectively, which indicates a high degree of accuracy. We note that these NRMSEs do not directly reflect the accuracy for $\eta_1$ and $\eta_2$ as each NRMSE is obtained based on five separate emulators (each of which is trained on a subset of the data). However, we expect both $\eta_1$ and $\eta_2$ to have roughly the same or better accuracy than the calculated CV accuracy since they learn from more data.

Since $\eta_1$ and $\eta_2$ are LMGPs, they provide an additional validation metric in the form of the latent spaces for both fidelity level and orientation. As shown in Figures 6a and 6b, both emulators learn that the mid-fidelity structures are more accurate than the low-fidelity structures and place the latent point for $y_m$ closer to $y_h$ than the latent point for $y_l$. The emulators also learn the expected behavior with regards to orientation—as shown in Figures 6c and 6d, the latent points for all three orientations are equidistant in a triangle pattern which indicates distinct uncorrelated behavior for each. We expect this behavior because we trained both emulators on a combined dataset containing both *Cyl*-type and *Sph*-type structures. If we were to instead separately train emulators for each, we would expect to see the latent points for $\mathscr{O}_1$ and $\mathscr{O}_2$ coincide for the LMGPs trained on *Sph*-type structures.

For $\eta_3$, we generate a dataset of size $500$ (using the same ranges for the design variables) and use $350$ and $150$ samples to, respectively, train and test the emulator. $\eta_3$ achieves an NRMSE of $0.218$ and a corresponding mean absolute error (MAE) of $2.400$,[17] indicating a reasonable degree of accuracy.

### 4.2 Multi-Objective Optimization Results

We demonstrate the results of our multi-objective optimization via GA in Figure 7. All six Pareto frontiers show the expected permeability-conductivity trade-off between $f_1$ and $f_2$. As shown in Figure 7a, the best performing combination of orientation and SDF type is $(\mathscr{O}_2, Cyl)$, which exceeds the performance of every other combination throughout the domain. We also note that $\mathscr{O}_3$, which corresponds to placing the microstructure such that its $x$ axis is aligned with the $z'$ axis of the wick system, performs much worse than the other orientations. We explain this by noting that our SDFs are constructed such that there exist nonzero points on the $x$ axis regardless of the choice of parameters This construction translates to frequency variation in the $x$ direction in the microstructure ($z'$ direction in the thermofluidic system). Hence, placing structures in orientation 3 (i.e., with the SDF $x$ axis oriented vertically in the wick), never results in continuous solid $z'$-direction connections, e.g., vertical sheets or pillars. This leads to lower maximum $z'$-direction conductivity and hence lower $f_2$, which we also see reflected in Figure 7. Additionally, we note that ($\mathcal{O}_1$, *Sph*) and ($\mathcal{O}_2$, *Sph*) have near-identical performance, which indicates that the GA results are performing as expected since as explained in Section 3.2.3, these two orientations are equivalent. Finally, we note that the value of $f_3$, shown in the color bar of Figure 7b, correlates positively with $f_1$ and negatively with $f_2$, i.e., , the number of solid clusters increases as permeability increases and as conductivity decreases. This trend is expected since both increasing permeability and decreasing conductivity correlate with decreasing volume fraction which leads to less well-connected quasi-random microstructures.

---
[17]We provide MAE for $\eta_3$ as it is directly interpretable as the average number of solid clusters.

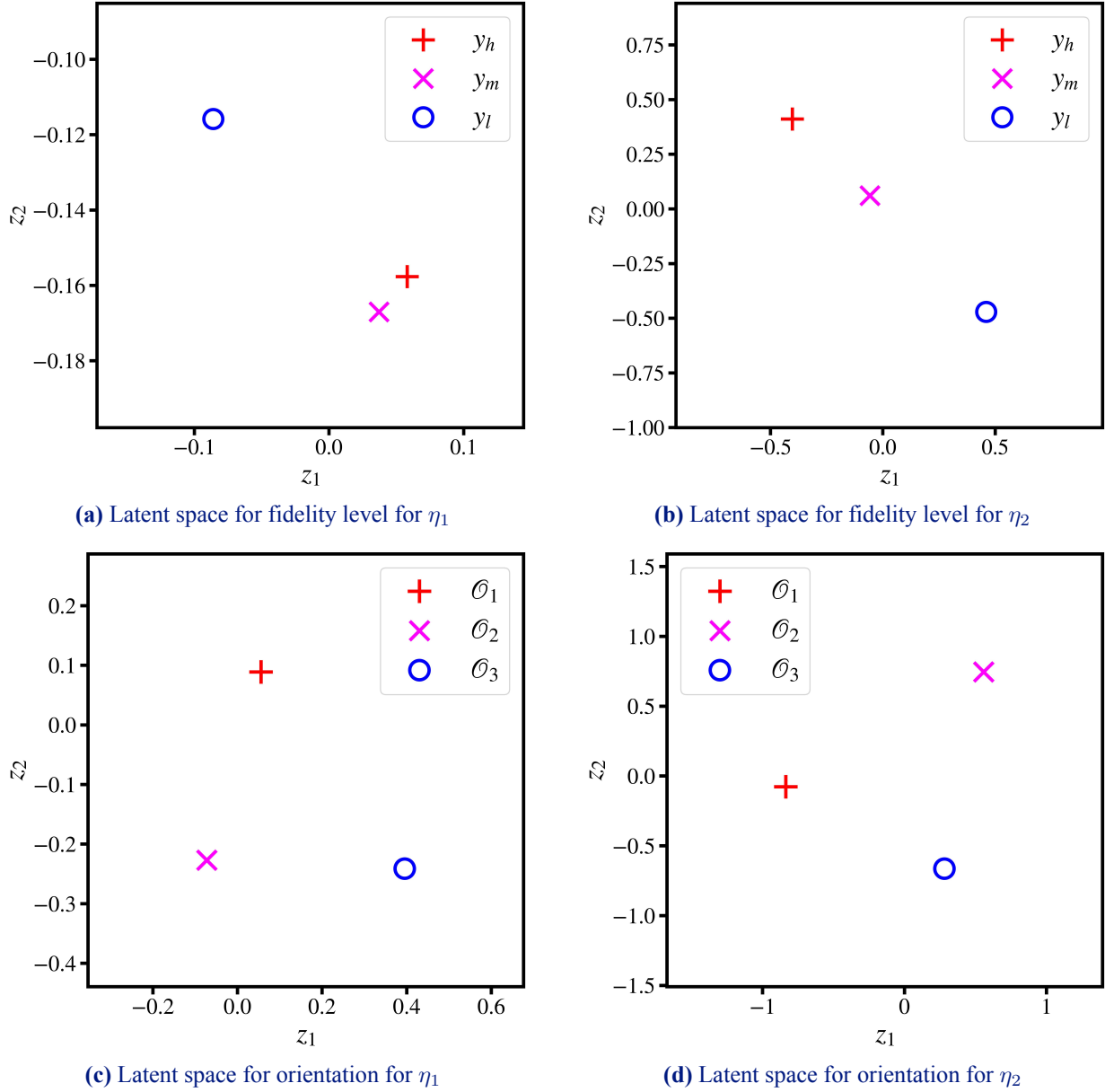

**Figure 6 Latent spaces learned by** $\eta_1$ **and** $\eta_2$**:** (a) $y_h$ and $y_m$ are close together in the latent space while $y_l$ is more distant, indicating that the low-fidelity samples are significantly less accurate than the mid-fidelity samples. b $y_h$, $y_m$, and $y_l$ lie on a line ordered by fidelity level. (c-d) $\mathcal{O}_1$, $\mathcal{O}_2$, and $\mathcal{O}_3$ are roughly equidistant, indicating that all three orientations display unique behaviors.

## 4.3 Optimized Structures

In this subsection, we first visualize some selected structures from the Pareto frontier in Section 4.3.1 and then analyze the performance of the structures in terms of the two objective functions in Section 4.3.2.

### 4.3.1 Visualization and Feasibility Assessment

We visualize seven optimized porous microstructures ordered by increasing $f_1$ (or decreasing $f_2$) selected from the best-performing Pareto frontier in Figure 8. As we progress from $\mathcal{M}_1$ to $\mathcal{M}_7$, we observe an evolution of structure topologies that demonstrates a decrease in solid volume fraction, characteristic frequency, and complexity. This aligns with our expectation that solid volume fraction correlates positively with $f_2$ (based on the thermal conductivity) and negatively with $f_1$ (based on the permeability). The decrease in high-frequency behavior also aligns with our expectations since high-frequency behavior leads to thinner channels and lower mass flow rates (and hence lower $f_1$).

As the structures range from $\mathcal{M}_1$ to $\mathcal{M}_7$, they become more well-ordered (less complex) and show

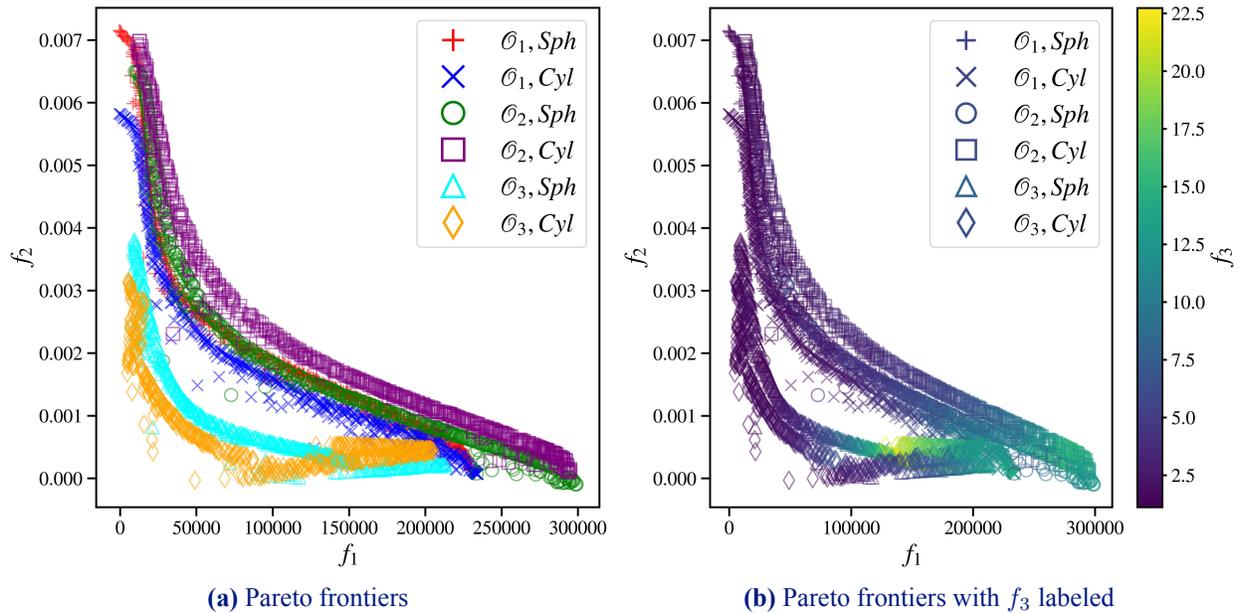

**(a)** Pareto frontiers      **(b)** Pareto frontiers with $f_3$ labeled

**Figure 7 Comparison of Pareto frontiers:** 2-D projections of the results of our 3-objective optimization. **(a)** Cylinder-type SDF placed in orientation 2 outperforms all other combinations throughout the entire design space. **(b)** $f_3$ generally increases in value with increasing $f_1$ and decreasing $f_2$.

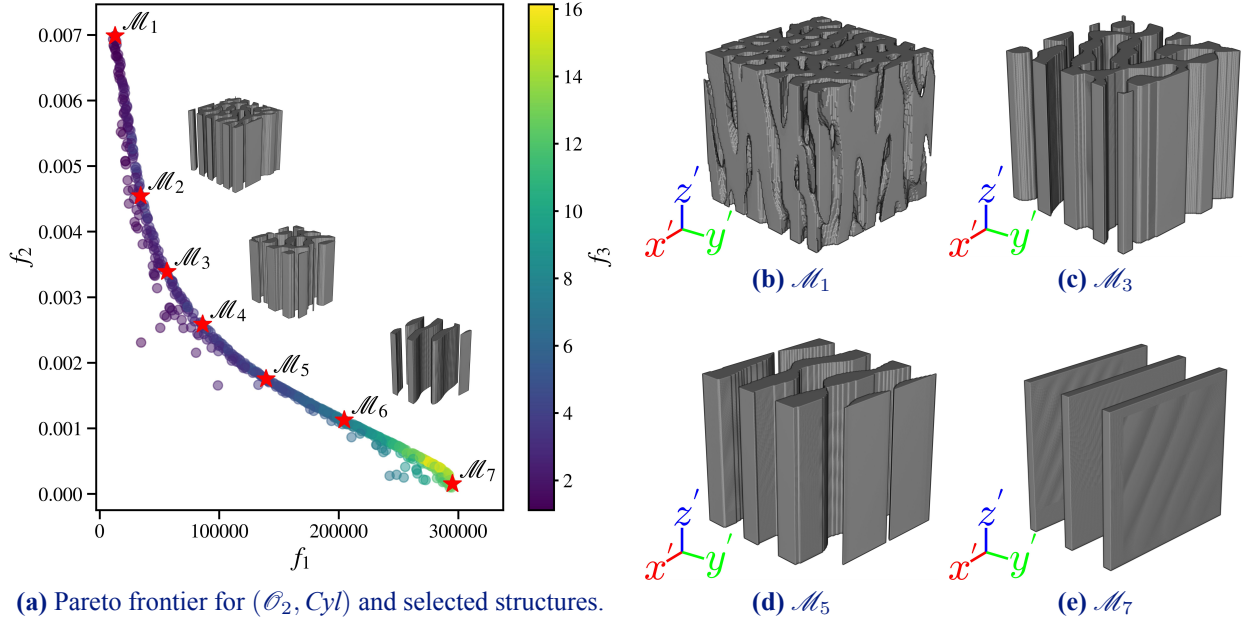

**(a)** Pareto frontier for $(\mathcal{O}_2, Cyl)$ and selected structures.   **(d)** $\mathcal{M}_5$   **(e)** $\mathcal{M}_7$

**Figure 8 Best performing Pareto frontier and selected microstructure visualizations: (a)** We select seven microstructures from the Pareto front to visualize. $\mathcal{M}_2$, $\mathcal{M}_4$, and $\mathcal{M}_6$ are shown next to their corresponding locations in the objective space while the other structures are described in **(b)-(e)**. **(b)** $\mathcal{M}_1$ shows behavior characteristic of the high-$f_2$ region with high volume fraction and no floating solid clusters. **(c)** Compared to $\mathcal{M}_1$, $\mathcal{M}_3$ has lower volume fraction and much straighter vertical walls. **(d)** $\mathcal{M}_5$ continues the trends of decreasing volume fraction and randomness, and is composed of a few wavy, thin vertical walls. **(e)** $\mathcal{M}_7$ is characteristic of the trends of structures seen in thermal management and the cooling industry—the microstructure is composed of three parallel, perfectly straight walls.

a concomitant decrease in the characteristic frequency of phase variation along each axis as well as the number of axes with significant variation. For example, in $\mathcal{M}_1$, we observe high-frequency phase variation in the $x'$ and $y'$ directions and low-frequency phase variation in the $z'$ direction. With $\mathcal{M}_3$, we lose all variation in the $z'$ direction and have lower-frequency variations in the $x'$ and $y'$ directions. $\mathcal{M}_5$ still has no variation in the $z'$ direction and has very low-frequency variation in the $x'$ direction with lower-frequency variation in the $y'$ direction than $\mathcal{M}_3$. Lastly, $\mathcal{M}_7$ only has phase variation in the $y'$ direction. While these trends are difficult to quantify in the microstructure space, they correspond to decreasing $r$, $\sigma$, $\theta$, and $\phi$ in the SDF inputs as we increase the structure index. These trends in the SDF parameters explain the previous two observed trends—$r$ directly correlates positively with the characteristic frequency of the phase variation of the microstructure (and hence negatively with the channel widths), while $\theta$ and $\phi$ control the degree of frequency variation along the $y$ and $z$ axes, respectively. $\sigma$ has a more subtle effect, with larger values causing a higher range of frequency behaviors throughout the microstructure which corresponds to higher variation in channel widths and hence more complex structures.

The number of distinct solid clusters is not strictly increasing as we go from $\mathcal{M}_1$ to $\mathcal{M}_7$—instead, $\mathcal{M}_1$ is fully connected while $\mathcal{M}_2$ through $\mathcal{M}_7$ are not. This aligns with our expectation that the *probability* of having more than one distinct solid cluster should correlate with decreasing $v$. Additionally, structures with phase variation in fewer directions are more likely to have disconnected clusters. For example, a structure built from the same SDF as $\mathcal{M}_7$ but with a higher volume fraction will still be composed of distinct, albeit thicker, disconnected sheets. While $\mathcal{M}_2$ through $\mathcal{M}_7$ are not fully connected, each distinct solid cluster is connected to the base of the design system (bottom $z'$-axis face) and therefore these designs are feasible.

Finally, we note that a fin-like topology in the out-of-plane direction emerges in our Pareto-optimal structures as a preferable configuration for cooling our thermofluidic system. A similar trend is seen in designs for commercial air- or liquid-cooling heat sinks [29, 30], which incorporate porous screen-fins and vertical channel-like structures, and these designs bear a striking similarity to the selected structures shown in Figure 8. However, these trends are not imbued into our design space or the search process; rather, they emerge from our fully-automated inverse design process as a natural result of the selected objectives in the multi-objective optimization step.

### 4.3.2 Visualizations and Validations of Thermofluidic Objectives

Our first objective controls the evaporative heat flux at the top surface of the microstructure and critically depends on the pore topology. Our second objective regulates the temperature uniformity in the $z'$ direction of the porous microstructure and is exclusively affected by the solid phase. To gain a deeper understanding on how these two objectives spatially vary across our Pareto optimum designs, we examine the fluid dynamics and thermal characteristics.

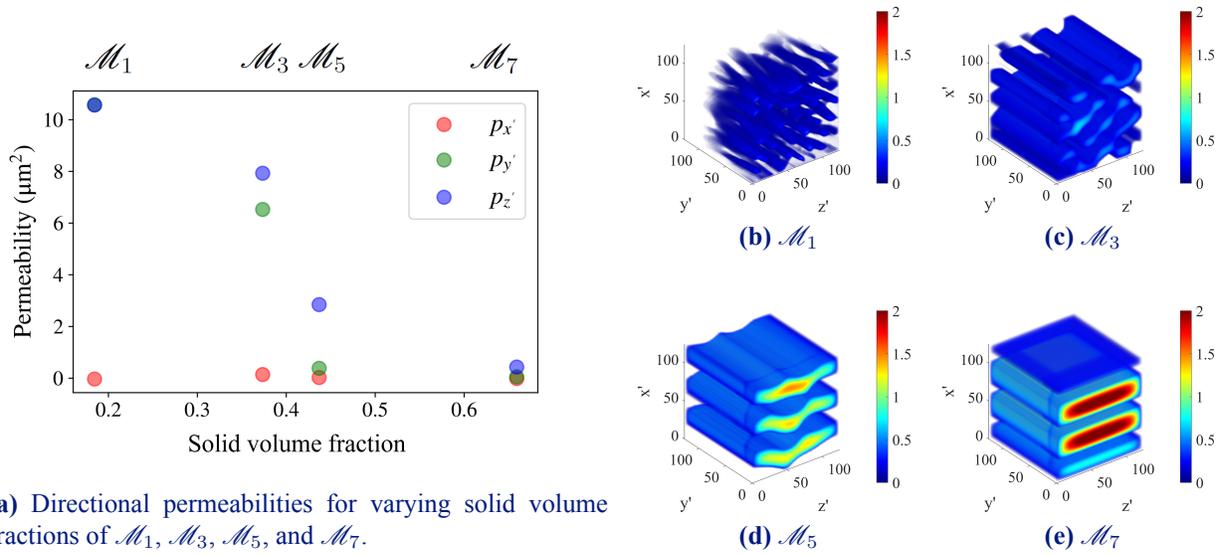

**(a)** Directional permeabilities for varying solid volume fractions of $\mathcal{M}_1$, $\mathcal{M}_3$, $\mathcal{M}_5$, and $\mathcal{M}_7$.

**(b)** $\mathcal{M}_1$   **(c)** $\mathcal{M}_3$

**(d)** $\mathcal{M}_5$   **(e)** $\mathcal{M}_7$

**Figure 9 Mass transfer behaviors for selected structures:** **(a)** Directional permeabilities ($p_{x'}$, $p_{y'}$, $p_{z'}$) extracted from LBM for varying solid volume fractions of the four selected optimal points on the Pareto frontier of objective space obtained from LMGP. **(b)-(e)** 3-D visualizations of the velocity fields in optimized porous structures of $\mathcal{M}_1$, $\mathcal{M}_3$, $\mathcal{M}_5$, and $\mathcal{M}_7$, respectively.

As elucidated in Section 3.2.3, the permeabilities in $x'$ and $y'$ directions dominantly contribute to the first objective $f_1$. As the structures progress from $\mathcal{M}_1$ to $\mathcal{M}_7$, they display noticeable declines in directional permeabilities across all axes, including $x'$, $y'$, and $z'$ directions, see Figure 9a. To visualize these trends in more detail, we provide four steady-state velocity fields corresponding to $\mathcal{M}_1$, $\mathcal{M}_3$, $\mathcal{M}_5$, and $\mathcal{M}_7$, see Figures 9b to 9e. In this context, a system reaches steady-state when the fluctuations in directional permeabilities, computed over a series of $10,000$ iterations, fall below a predetermined threshold value of $0.0005\ \mu m^2$. These velocity fields indicate that as the topological complexity increases in a given direction, it disrupts the formation of a concentrated flow path or primary strain. This disruption leads to a more dispersed flow field and consequently adversely affects the permeability in that specific direction of the microstructure.

Our second objective $f_2$ is primarily governed by the thermal conductivity in $z'$ direction, as explained in Section 3.2.3. Predictably, transitioning from $\mathcal{M}_1$ to $\mathcal{M}_7$, there is a discernible increase in the directional thermal conductivities along all axes, including $x'$, $y'$, and $z'$ directions, see Figure 10a. Here, $k_{max}$ represents the maximal theoretical thermal conductivity attainable by a microstructure, given a specified solid volume fraction. To study the correlation between directional thermal conductivities and solid topology, we visualize four temperature fields on the basis of PNMs for $\mathcal{M}_1$, $\mathcal{M}_3$, $\mathcal{M}_5$, and $\mathcal{M}_7$, respectively, see Figures 10b to 10e. The observed temperature fields suggest that the magnitude of directional thermal conductivity is primarily tied to the solid volume fraction of a microstructure. As the significance of $f_2$ escalates, denser PNMs emerge as more preferable.

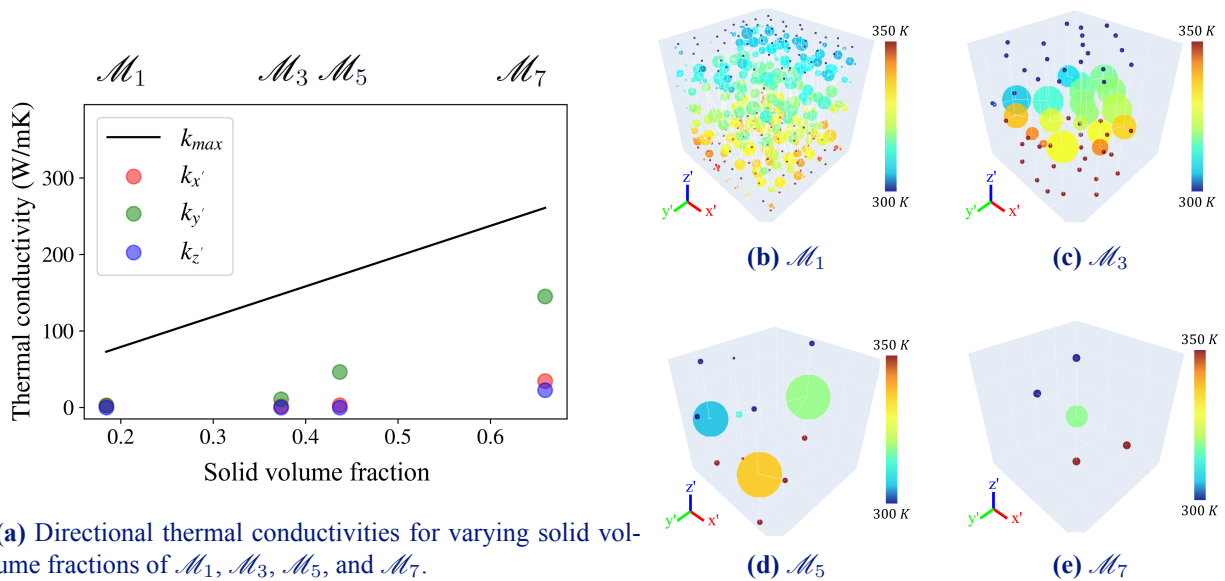

(a) Directional thermal conductivities for varying solid volume fractions of $\mathcal{M}_1$, $\mathcal{M}_3$, $\mathcal{M}_5$, and $\mathcal{M}_7$.

(b) $\mathcal{M}_1$

(c) $\mathcal{M}_3$

(d) $\mathcal{M}_5$

(e) $\mathcal{M}_7$

**Figure 10 Heat transfer behaviors for selected structures:** **(a)** Directional thermal conductivities ($k_{x'}$, $k_{y'}$, $k_{z'}$) extracted from PNMs for varying solid volume fractions of the four selected optimal points on the Pareto frontier of objective space obtained from LMGP, along with the upper limit thermal conductivity ($k_{max}$) in a porous structure. **(b)-(e)** 3-D visualizations of the PNMs and temperature fields in optimized porous structures of $\mathcal{M}_1$, $\mathcal{M}_3$, $\mathcal{M}_5$, and $\mathcal{M}_7$, respectively.

Finally, we validate our selected Pareto-optimal microstructure designs by extracting thermofluidic properties (according to the process described in Section 3.2) and comparing the objectives calculated from those properties to those predicted by our multi-fidelity LMGP emulators. The results, as shown in Table 2, indicate that both emulators $\eta_1$ and $\eta_2$ perform reasonably accurately on our test samples. However, $\eta_1$ loses accuracy in the high-$f_1$ regime, as shown in Table 2a, and $\eta_2$ develops a slight bias in the low-$f_2$ regime. We explain this by noting that the distribution of outputs in our training dataset has few samples with high-$f_1$ or low-$f_2$, making it difficult for our emulators to accurately predict the behavior of the corresponding structures. This comes from our intentional omission of structures with floating solid clusters during our data generation process (see Section 3.3) which encourages our dataset to only cover the space of physically feasible structures. Structures with floating solids occur at a much higher rate for lower volume fractions, leading to samples with high $f_1$ and/or low $f_2$ being disproportionately scarce in the dataset[18].

**Table 2 Comparison of emulator predictions and extracted properties for selected structures: (a)** Emulator predictions are accurate for the low-$f_1$ regime, but lose accuracy in the high-$f_1$ regime. **(b)** Emulator predictions are reasonably accurate throughout.

| Structure No. | $f_1$ simulated | $f_1$ emulated |
|---|---|---|
| $\mathcal{M}_1$ | 11060 | 12834 |
| $\mathcal{M}_2$ | 37386 | 34156 |
| $\mathcal{M}_3$ | 60804 | 56318 |
| $\mathcal{M}_4$ | 75422 | 86116 |
| $\mathcal{M}_5$ | 125221 | 139183 |
| $\mathcal{M}_6$ | 140260 | 204533 |
| $\mathcal{M}_7$ | 182182 | 295024 |

(a) $f_1$ validation

| Structure No. | $f_2$ simulated | $f_2$ emulated |
|---|---|---|
| $\mathcal{M}_1$ | 0.007256 | 0.006985 |
| $\mathcal{M}_2$ | 0.003471 | 0.004545 |
| $\mathcal{M}_3$ | 0.002322 | 0.003390 |
| $\mathcal{M}_4$ | 0.001485 | 0.002587 |
| $\mathcal{M}_5$ | 0.000546 | 0.001753 |
| $\mathcal{M}_6$ | 0.000549 | 0.001127 |
| $\mathcal{M}_7$ | 0.000137 | 0.001531 |

(b) $f_2$ validation

## 5 Conclusion

We develop a data-driven framework for automatically and systematically designing quasi-random porous microstructures for microelectronic cooling. Our framework reduces the design space to tractable dimensionality by leveraging SDFs and balances cost and accuracy through offline multi-fidelity simulations. It quantifies the performance and feasibility of design candidates through property-based objectives, builds accurate multi-fidelity emulators providing visualizations of relationships within the datasets, and uses these emulators to find Pareto-optimal design candidates. We demonstrate and validate our framework by applying it to the design of porous microstructures for thin-film evaporation where our results reflect the well-established trends seen in the industry.

To reduce the dimensionality of the problem, we develop two classes of parameterized SDFs which encode our design space via a small number of meaningful parameters and demonstrate that this enables us to explore a wide variety of microstructures. The results of the multi-objective optimization demonstrate that the cylinder-type parameterization placed in orientation 2 solidly outperforms the other combinations of SDF type and orientation. We also observe, as expected, identical behavior from two orientations of the sphere-type parameterizations and the trend that expected number of solid clusters correlates negatively with volume fraction. These observations further cement that our emulators are well-trained. The structures selected from the Pareto front also align with our prior expectations. Interestingly, these designs, which are found fully automatically via inverse design, align with heat sink design in industry applications despite no preference for these designs being imbued in our optimization process. Despite containing multiple solid clusters, our selected designs are feasible as they connect to the base of the system. Concurrently, the chosen microstructures exhibit notable fluid transport in the $x'$ and $y'$ directions, along with efficient thermal delivery in the $z'$ direction, meeting the cooling demands of the target thermofluidic system in this work.

---

[18]We plan to address this in future works by developing methods to directly generate fully-connected / feasible structures.

The framework we have developed facilitates rapid, agile, and systematic searching for optimal microstructures, and can be easily adapted to other problems. However, our framework as presented in this paper is limited to homogeneous microstructure designs as it is based off of encoding the design space via SDFs which cannot reconstruct nonstationary microstructures. Given that such designs often outperform homogeneous ones, especially when the heat flux into the system is not uniform, we plan to extend our framework to accommodate heterogeneous microstructure design. We also plan to explore designs using other engineered surfaces such as micropillars.